\documentclass[10pt,twocolumn,letterpaper]{article}

\usepackage{iccv}
\usepackage{times}
\usepackage{epsfig}
\usepackage{graphicx}
\usepackage{amsmath}
\usepackage{amssymb}
\usepackage{multirow}
\usepackage[misc]{ifsym}
\usepackage[accsupp]{axessibility}  

\usepackage[breaklinks=true,bookmarks=false]{hyperref}

\iccvfinalcopy 

\def\iccvPaperID{****} 

\ificcvfinal\fi

\begin{document}

\title{Spatial Uncertainty-Aware Semi-Supervised Crowd Counting}

\author{Yanda Meng\textsuperscript{1}, Hongrun Zhang\textsuperscript{1}, Yitian Zhao\textsuperscript{2}, Xiaoyun Yang\textsuperscript{3}, Xuesheng Qian\textsuperscript{4}, Xiaowei Huang\textsuperscript{5}, \\   Yalin Zheng\textsuperscript{1} \Letter
\\ yalin.zheng@liverpool.ac.uk
\\ \textsuperscript{1} Department of Eye and Vision Science, University of Liverpool, Liverpool, United Kingdom
\\ \textsuperscript{2} Cixi Institute of Biomedical Engineering, Chinese Academy of Sciences, Ningbo, China
\\ \textsuperscript{3} Remark AI UK Limited, London, United Kingdom
\\ \textsuperscript{4} China Science IntelliCloud Technology Co., Ltd, Shanghai, China
\\ \textsuperscript{5} Department of Computer Science, University of Liverpool, Liverpool, United Kingdom

}

\maketitle
\ificcvfinal\thispagestyle{empty}\fi

\begin{abstract}
Semi-supervised approaches for crowd counting attract attention, as the fully supervised paradigm is expensive and laborious due to its request for a large number of images of dense crowd scenarios and their annotations.
This paper proposes a spatial uncertainty-aware semi-supervised approach via regularized surrogate task (binary segmentation) for crowd counting problems. 
Different from existing semi-supervised learning-based crowd counting methods, to exploit the unlabeled data, our proposed spatial uncertainty-aware teacher-student framework focuses on high confident regions' information while addressing the noisy supervision from the unlabeled data in an end-to-end manner.
Specifically, we estimate the spatial uncertainty maps from the teacher model's surrogate task to guide the feature learning of the main task (density regression) and the surrogate task of the student model at the same time. 
Besides, we introduce a simple yet effective differential transformation layer to enforce the inherent spatial consistency regularization between the main task and the surrogate task in the student model, which helps the surrogate task to yield more reliable predictions and generates high-quality uncertainty maps.
Thus, our model can also address the task-level perturbation problems that occur spatial inconsistency between the primary and surrogate tasks in the student model. 
Experimental results on four challenging crowd counting datasets demonstrate that our method achieves superior performance to the state-of-the-art semi-supervised methods. 
Code is available at : \url{https://github.com/smallmax00/SUA_crowd_counting}
\let\thefootnote\relax\footnotetext{Yalin Zheng \textsuperscript{1} is the corresponding author.}
\end{abstract}

\section{Introduction}\label{sec:introduction}

The task of crowd counting in computer vision is to infer the number of people in images or videos. There is an ever-increasing demand for automated crowd counting techniques in various applications such as public safety, security alerts, transport management \(etc. \).

With the help of Convolutional Neural Network (CNN)'s feature learning ability, current state-of-the-art methods \cite{bai2020adaptive,yang2020reverse,wan2019residual,zhang2019wide,sindagi2019multi,shi2018crowd} gained excellent counting performance by regressing the corresponding density maps of the input images, where the summed value in a density map gives the total count numbers.
To train a robust and accurate crowd counting estimator, most of the existing methods \cite{liu2019context,shi2019revisiting,liu2019point,liu2019adcrowdnet,jiang2019crowd,shen2018crowd} relied on substantial labeled images, where head centres must be annotated for training. However, the annotation process can be labour-intensive and time-consuming. For example, JHU-Crowd \cite{sindagi2020jhu} dataset contains labels of 1.51 millions people whilst NWPU-Crowd \cite{gao2020nwpu} dataset contains annotations of 2.13 millions people, which takes 3000 human hours in total. Hence, reducing annotation efforts while maintaining good counting performance is our goal in this paper. More specifically, we study the counting estimator in a semi-supervised manner where limited labeled data is used; on the other hand, the unlabeled data is leveraged to improve our model's robustness and performance. 

\begin{figure*}
\centering
\includegraphics[width=18cm]{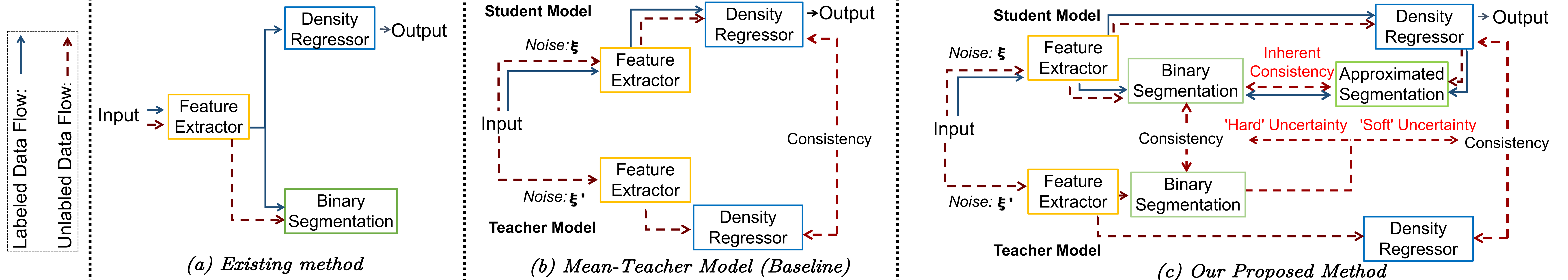}
\caption{Overview of the very recent work \cite{liu2020semi}, baseline model \cite{tarvainen2017mean}, and our proposed method. (a): \cite{liu2020semi} utilized surrogate task (binary segmentation) to boost the feature extractor with labeled and unlabeled data so as to enhance the performance of the density regressor.
(b): Mean-Teacher \cite{tarvainen2017mean} is a commonly used semi-supervised framework through exploiting the consistency learning on the student and the teacher models' outputs under different input-level noise perturbations \((\xi, \xi{'})\) and model-level noise perturbation (Dropout \cite{srivastava2014dropout} of the student and teacher models).
We refer to it as the baseline model in this paper. 
(c): Our Mean-Teacher based semi-supervised framework. Note that we only input the unlabeled data into the teacher model because this work aims to explore the unlabeled data's uncertainty. The estimated `hard' and `soft' spatial uncertainty maps aim to assist the consistency learning (upon binary segmentation and density regression) between the student and teacher models; one can alleviate the unlabeled data's inevitable noisy supervision. The student model's binary segmentation is regularized by the inherent consistency regularization with approximated segmentation to address the spatial predictions' perturbation issues between binary segmentation and density regression tasks in the student model. } 
\label{overview}
\end{figure*}
Previous semi-supervised crowd counting methods tend to minimize the expensive label work through active learning \cite{zhao2020active,liu2019exploiting}, synthetic images \cite{wang2019learning,wang2020pixel}, or pseudo-ground truth \cite{sindagi2020learning,liu2020semi}. However, they did not consider the unlabeled data or synthetic data's intrinsic noisy supervision due to the inherent data uncertainties \cite{oh2020crowd}. 
Uncertainty estimation has been explored in other computer vision tasks, such as segmentation \cite{kohl2018probabilistic,yu2019uncertainty,baumgartner2019phiseg} or detection \cite{Zhang2020UCNet,he2019bounding}, \(\etc.\).     
There are two significant types of uncertainty \cite{kendall2017uncertainties}: epistemic uncertainty, which accounts for the uncertainty in the model parameters and can be addressed when given enough data; aleatoric uncertainty corresponds to inevitable noisy perturbation existing in the data itself. Solving the aleatoric uncertainty is a crucial problem since crowd images contain inherent noises such as complex backgrounds, massive occlusions and illumination variations \(etc.\).
Few recent approaches \cite{oh2020crowd,ranjan2020uncertainty} have considered the uncertainty quantification in the crowd counting task in a fully-supervised manner. They adopted \cite{kendall2017uncertainties} to estimate the mean and variance of the assumed Gaussian distribution of the density map, where the variance is served as a measure of uncertainty. 

In this work, we exploit the aleatoric uncertainty in a semi-supervised manner to alleviate the noisy supervision in uncertain spatial regions due to the complex backgrounds and massive occlusions challenges from the unlabeled crowd images \cite{oh2020crowd}. 
Previous crowd counting methods \cite{zhao2019leveraging,shi2019counting,gao2019pcc} prove that the spatial region information from the binary segmentation task is essential to tell the crowd and background locations, which will help the density map regressor to focus on the region of interest and improve the counting performance. In our work, the binary segmentation provides spatial information and serves as a surrogate task to estimate the uncertain spatial regions (\(e.g.\) uncertain crowd locations). 
With the estimated spatial uncertainty, we assist the unsupervised consistency learning (upon binary segmentation and density regression) between the student model and the teacher model based on the Mean-Teacher \cite{tarvainen2017mean} semi-supervised learning framework. Fig. \ref{overview} (b \& c) shows the overview structure of our method and the re-implemented Mean-Teacher framework \cite{tarvainen2017mean} for the crowd counting task. Note that, in our work, the student model and the teacher model share a similar structure (Feature extractor, binary segmentation module, density regressor). We update the teacher model's parameters as an exponential moving average \((EMA)\) of the student model's parameters. Because ensembling the student model's predictions at different training steps can enhance the performance of the teacher model's predictions \cite{laine2016temporal}; in which case, the teacher model can generate `targets' for the student model to learn from. However, as mentioned above, those `targets' contain spatial-wise uncertainty; thus, we purify the `targets' with the estimated `hard' and `soft' uncertainty map during training.



Apart from the aforementioned novel components, we also study how to learn an excellent surrogate task (binary segmentation) predictor to produce reliable and consistent spatial uncertainty that the main task (density regression) has in the student model. Note that, followed by \cite{tarvainen2017mean},  only student model is used for the inference process.
Specifically, we introduce a simple yet effective differentiable transformation layer to approximate the binary segmentation maps from the density map predictions of the unlabeled input in the student model. 
We then employ an unsupervised inherent consistency loss between the predicted segmentation maps and the approximated segmentation maps to guarantee the consistent spatial feature learning between two different tasks in the student model. 
The underlying motivations are twofold: 
(1) the surrogate and the main task may introduce an inherent prediction perturbation on spatial regions due to the domain gap of feature learning from multi-tasks \cite{luo2021semi}. Our ablation experiment results prove that this perturbation will lead to noisy supervision upon two tasks, thus reducing the performance. 
(2) The proposed transformation layer itself is simple. However, it brings several benefits with the inherent consistency loss. For example, the estimated uncertainty from a regularized surrogate task can indicate more reasonable and consistent spatial uncertain regions that the main task has, which further enhances the consistency between the surrogate and the main task. In other words, with the proposed transformation layer, the estimated uncertainty and consistency regularization can benefit from each other to advance the counting performance. Our experiment results demonstrate that the proposed consistency regularization mechanism can boost the model's performance in both supervised and semi-supervised manner.

In summary, this work makes the following contributions:
(1) We propose a surrogate task to estimate the uncertain spatial regions from the unlabeled data under the semi-supervised Teacher-Student framework, which can alleviate the inevitable noisy supervision from the unlabeled data.
(2) We propose a differentiable transformation layer that enables the inherent spatial consistency regularization between the surrogate task (binary segmentation) and the main task (density regression) in the student model, which
can enhance the model to estimate high-quality uncertainty maps from the unlabeled data, thus improve our model's counting performance.
(3) We conduct extensive experiments on four well-known challenging counting benchmarks. Quantitative results demonstrate that our methods outperform existing semi-supervised crowd counting methods. Besides, with less than half of the labeled data, our method can achieve comparable performance with the fully-supervised state-of-the-art methods.



\section{Related Works}
Deep Learning based works has achieved superior performance in many computer vision tasks, such as classification \cite{bridge2020introducing,zhang2021regularization}, segmentation \cite{meng2020regression,meng2020cnn,chen2019learning,zhao2015automated}, and registration \cite{chen2021unsupervised}. In this section, we will discuss and compare with deep-learning based crowd counting methods in different supervision manners.

\subsection{Supervised Density-based Crowd Counting}
Recently, fully-supervised density map regression-based counting methods with CNN achieved good performance. Approaches like \cite{boominathan2016crowdnet,zhang2016single,zhang2015cross} proposed a multi-column network to regress the density map in terms of combining local and global features to tackle the scale variation challenges. Other works \cite{miao2020shallow,jiang2020attention,zhang2019attentional} employed visual attention mechanisms to address other issues, such as background noise in crowded cluster scenarios and various density levels from scale variations. Apart from single-task learning, recent works introduced auxiliary task learning frameworks, i.e. classification \cite{shi2019counting,sindagi2019ha}, localization \cite{liu2018decidenet,sam2020locate,liu2019recurrent,lian2019density,luo2019hybrid}, or segmentation \cite{zhao2019leveraging,shi2019counting,gao2019pcc}, which attains additional spatial and semantic information supplement from the joint learning auxiliary tasks. 
The above methods focus on improving the counting performance in a fully-supervised paradigm. However, annotating the crowd counting dataset is labour-intensive and time-consuming work. In this paper, we made efforts on minimizing the expensive labelling work in a semi-supervised manner.

\subsection{Learn to count with limited data}
Relieving the crowd counting annotation efforts by using weakly/semi-/un-supervised learning mechanism has attracted researchers' attention for the past two years.
For example, Liu \textit{et al.} \cite{liu2018leveraging} leveraged a large number of unlabeled images and introduced a pairwise ranking loss to estimate the density map. Along the same line, Yang \textit{et al.} \cite{yangweakly2020} proposed a soft-label sorting network to regress the counting numbers rather than density map, which results in a performance reduction because of the difficult optimization from the input images to the target of scalar. Further, Wang \textit{et al.} \cite{wang2019learning,wang2020pixel} focused on a different direction, where they combined the synthetic images and realistic images to minimize the annotation burden. However, there is a domain gap between the synthetic and real-world scenarios; thus, they need further manual selections from the synthetic data. More recently, pseudo-labeling based semi-supervised approaches \cite{sindagi2020learning,liu2020semi} estimated the pseudo-ground truth of the unlabeled data, which is then used to supervise the network and improve the performance. Similarly, active learning-based methods \cite{zhao2020active,liu2019exploiting} annotated the most informative images instead of the whole training dataset and learned to count upon them. These methods can be effectively performed on the unlabeled data, but the model may be misled by the inevitable noisy supervision from the unlabeled data due to the aleatoric uncertainties \cite{oh2020crowd}, such as massive occlusions, complex backgrounds, \textit{etc}.


\subsection{Most Related Works}
The framework of the most recent state-of-the-art method \cite{liu2020semi} is shown in Fig. \ref{overview} (a), where the surrogate task (binary segmentation) learning mechanism is used to learn a robust feature extractor in a semi-supervised manner.
We believe that learning a better feature extractor can be more reliable towards the unlabeled data's noisy supervision. However, there are some fundamental limitations in their framework: 
(1) The unlabeled data are only used to train the feature extractor and the binary segmentation predictor, aiming to avoid noise from unlabeled data contaminating the density regressor. However, it also leads to a side effect that only limited labeled data is used to train the density map predictor, subject to sub-optimal results.
(2) Due to the unlabeled data's inevitable inherent noise, their model may provide incorrect predictions with spuriously high confidence because of the noisy supervision. This challenge has also been observed in other weakly/semi-/un-supervised crowd counting methods \cite{sam2019almost,von2016gaussian,sindagi2020learning,liu2018leveraging,yangweakly2020}. 
(3) The inherent prediction perturbation on spatial regions between the binary segmentation task and the density regression task may mislead the feature extractor's feature learning. In other words, the spatial inconsistency exists in the binary segmentation and density regression task. 

We propose a semi-supervised model to address all the limitations mentioned above, and a simplified diagram of the model is shown in Fig. \ref{overview}(c). 
Firstly, we introduced novel `hard' uncertainty and `soft' uncertainty from the teacher model to assist the student network to learn high-confident binary segmentation and density map predictions of the unlabeled data. This can alleviate the inevitable noisy supervision from the unlabeled dataset.
Secondly, we proposed a novel differentiable transformation layer that converts the predicted density maps into approximated binary segmentation maps, where the inherent consistency loss is employed to avoid the prediction perturbations issues. 
Thirdly, because of the proposed uncertainty map and inherent consistency regularization, the feature extractor, binary segmentation predictor and density regressor in the student model of our work can benefit from both the labeled and unlabeled data and avoid sub-optimal issues; details of the proposed components are explained in the following sections.

\section{Methods}

The ground truth of the density map is generated by \cite{lempitsky2010learning}. The binary segmentation ground truth mask is generated from the density map ground truth. Specifically, the value for each pixel in the binary ground truth mask is set to 1 if the pixel value of the density map is greater than 0, and 0 otherwise. 

\begin{figure*}
\centering
\includegraphics[width=12.6cm]{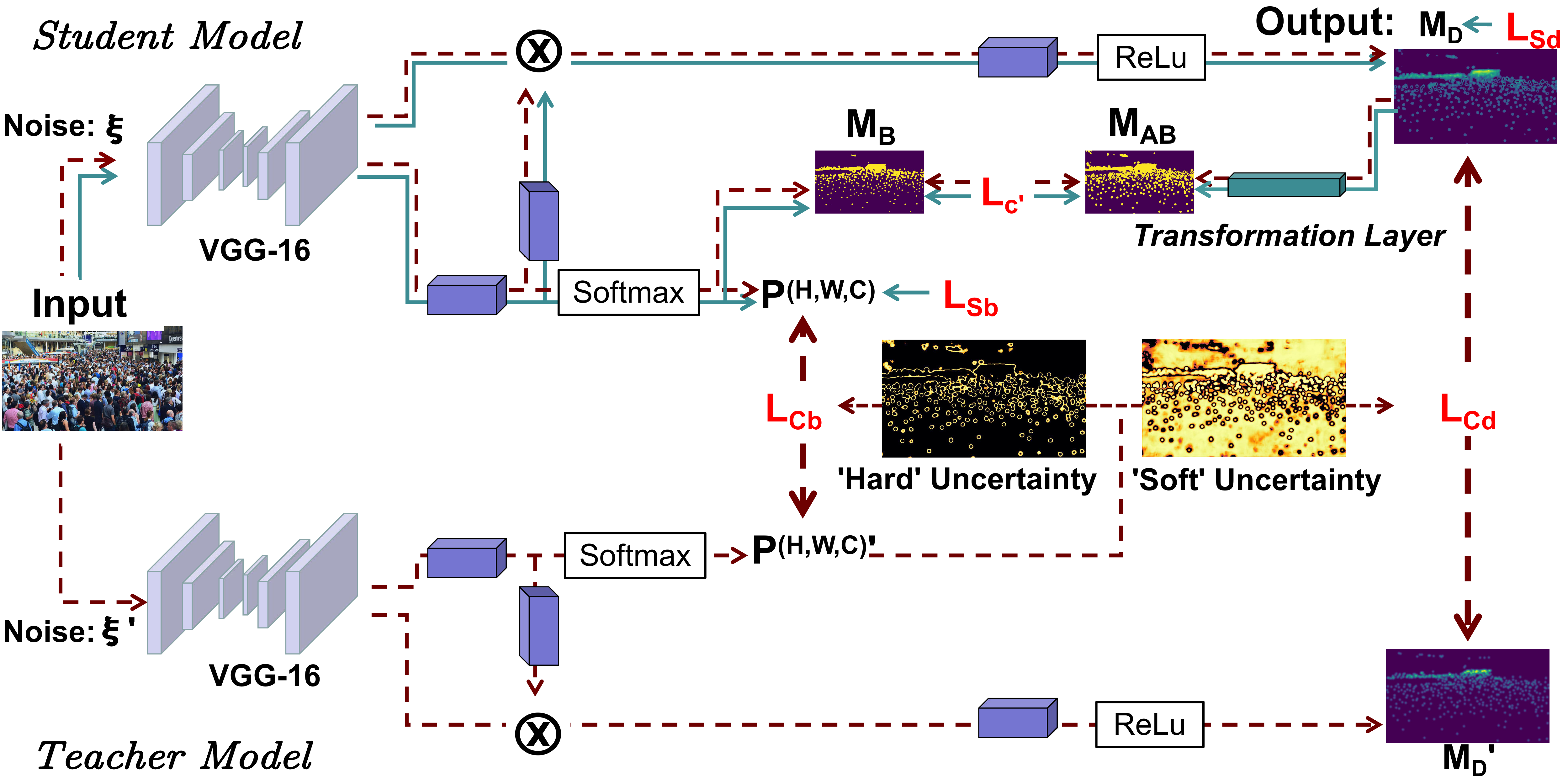}
\includegraphics[width=4.4cm]{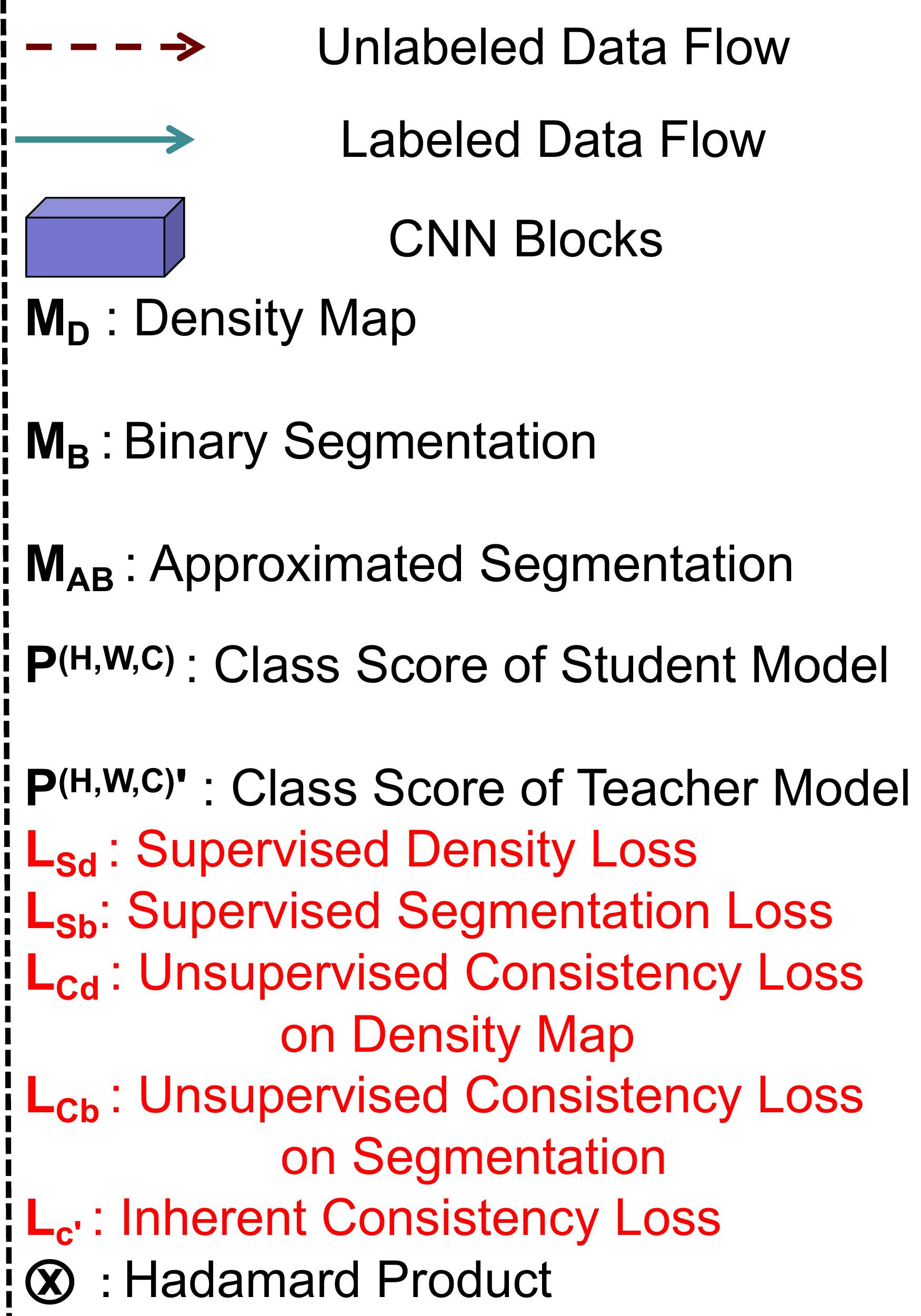}
\caption{The pipeline of our uncertainty-aware framework for semi-supervised crowd counting enabled by the regularized surrogate task. The student model is optimized by minimizing the supervised density regression loss \(L_{Sd}\), the binary segmentation loss \(L_{Sb}\) on the labeled data; the unsupervised inherent consistency loss \(L_{c{'}}\) on both the unlabeled data and labeled data, the unsupervised consistency loss \(L_{Cb}\) and \(L_{Cd}\) on the unlabeled data. The estimated spatial uncertainty (`hard' and `soft' uncertainty maps) from the teacher model guides the student to learn more reliable targets from the teacher. \(P^{(H,W,C)}{'}\) and \(M_{D}{'}\) are the outputs of the predicted class score and density map from the teacher model, which serves as the targets for the student model to learn from through consistency loss \(L_{Cb}\) and \(L_{Cd}\) respectively.}
\label{network}
\end{figure*}

The proposed Teacher-Student framework structure is illustrated in Fig. \ref{network}.
The uncertainty map is estimated from the surrogate task with unlabeled data in the teacher model. Then we use the uncertainty map to assist the surrogate and the main task feature learning in the student models.
The inherent consistency regularization between the surrogate task (binary segmentation) and the main task (density regression) in the student model improves its robustness regarding task-level spatial crowd region consistency. 

\subsection{Uncertainty Map Estimation}
Different from the recent fully-supervised Gaussian distribution uncertainty-based \cite{kendall2017uncertainties} crowd counting method \cite{ranjan2020uncertainty,oh2020crowd}, we propose a semi-supervised method to estimate the spatial uncertainty from the surrogate task (binary segmentation) in the teacher model with the unlabeled data, then use the uncertainty to assist the binary segmentation and density regression tasks feature learning in the student model so as to address the noisy supervision. This design is motivated by three considerations: 
(1) For crowd counting, the inevitable noise exists in many scenes, such as massive occlusions, complex backgrounds, \(etc.\), which results in uncertain crowd regions \cite{oh2020crowd}. So, the guidance of the proposed spatial uncertainty from the binary segmentation can be essential to alleviate the effects of noise.
(2) Without the annotations in the unlabeled inputs, the predicted outputs from the teacher model may be unreliable and noisy. Therefore, an uncertainty-aware learning scheme is essential for the student model to assess the uncertainty and conduct a more reliable consistent feature learning.
(3) The uncertainty estimated from the binary segmentation task indicates the uncertain locations of the crowd, which should be considered in the density regression task. Because the non-crowd regions should only maintain zero pixel values in the density map, the density regressor may produce larger pixel values due to the unlabeled data's spatial noise.

Recent domain adaptation studies \cite{long2018transferable,vu2019advent,zou2019confidence} indicated that due to the domain gap, the models trained on source domain tend to produce under-confident, \textit{i.e.} high-entropy predictions on the target domain.
We found that such a phenomenon also exists in semi-supervised crowd counting tasks. Specifically, in our model, the outputs of the binary segmentation with unlabeled data in the teacher model tend to produce under-confident regions (the boundary along crowd regions). As mentioned in \textit{Section.} \ref{sec:introduction} , this is because of the inevitable noise of the unlabeled data. Please refer to Fig. \ref{allresults} for the qualitative uncertainty visualisation. To address this challenge, we adopt Shannon Entropy \cite{shannon1949mathematical} as the metric to measure the randomness of the information \cite{shannon2001mathematical}, which is referred to as the uncertainty in this work. We then propose the `hard' and `soft' uncertainty maps to purify the learning process with the unlabeled data.
Formally, given a \(C\)-dimensional softmax predicted class score \(P_{x}^{(H,W,C)}\) from a \(H \times W\) input image \(x\), the Shannon Entropy \(I_{x}^{(H,W)}\) is defined as: 
\begin{equation}
    \label{selfinformation}
    I_{x}^{(H,W)} = -\sum_{c=1}^{C}P_{x}^{(H,W,C)} \odot \log{P_{x}^{(H,W,C)}},
\end{equation}
where \(\odot\) is Hadamard Product; \(C\) is the number of classes, which is 2 in our work because of the binary segmentation. 
In practice, we perform \(T\) times stochastic forward passes on the teacher model under random dropout and Gaussian noise input for each unlabeled input image. Therefore, we obtain a set of softmax probability vectors: \(\bigl\{P^{t}\bigr\}_{t=1}^{T}\) from the segmentation branch, then the predicted class score \(P^{(H,W,C)}\) is equal to \(\frac{1}{T} \sum_{t=1}^{T}P^{t}\), thus we can obtain \(I^{(H,W)}\) with equation \ref{selfinformation}. 

With the assistance of the approximated Shannon Entropy \(I^{(H, W)}\), we design two strategies to address the spatial uncertainty upon binary segmentation and density regression tasks between the student model and the teacher model, respectively.
Firstly, the `hard' uncertainty map \(U_{h}\) is introduced to guide the consistency learning on binary segmentation. In detail, we set a \textit{threshold} and filter out the relatively unreliable binary segmentation predictions of the teacher model and select only the certain predictions as the target for the student model to learn from. In practice, the `hard' uncertainty map \(U_{h}\) is equal to \( 1(I_{H,W} < threshold)\), where \(1(\cdot)\) is an indicator function.
Secondly, the `soft' uncertainty map \(U_{s}\) is proposed to assist the consistency learning on the density regression task. The uncertain crowd regions do contain noisy density map predictions; however, only relying on the spatial uncertainty and filtering out the uncertain density map predictions may mislead the density regression or even add more noises. In addition to the spatial uncertainty, there are also other uncertainties caused by perspective distortions, non-uniform distribution, weather changes, \(etc.\). Our ablation study experiments prove that the `soft' uncertainty maps are more friendly than `hard' uncertainty maps regarding to the consistency learning upon density map regression, thus advancing a performance boost.  
We retain all the density map predictions of the teacher model and introduce the `soft' uncertainty map as a weighted mask to assign different weights to each pixel on the density map prediction according to the spatial certainty level. In detail, for these relatively reliable regions (pixels), we assign them more weights during the training to enforce the consistent learning to focus on the certain prediction regions, while the relatively uncertain regions are still involved during the training with lower weights. In practice, we normalize the estimated Shannon Entropy \(I^{(H,W)}\) into range \((0,1)\) as  \(\hat{I}^{(H,W)}\), then define the `soft' uncertainty map \(U_{s}\) as: \(U_{s} = M * (1 - \hat{I}^{(H,W)})\) , where \(M\) is the constant value of weighted mask to control the \(U_{s}\) pixel values.

With the estimated `hard' and `soft' uncertainty map, the uncertainty-aware unsupervised consistency loss (\(L_{Cd}\) \& \(L_{Cb}\)) upon the main task (density regression) and surrogate task (binary segmentation) between the teacher and the student models can be guided during the training. Details will be shown in \textit{Section.} \ref{sec:lossfunction}.

\subsection{Regularized Surrogate Task via Transformation Layer}
In Fig. \ref{overview} (a), recent method \cite{liu2020semi} utilized surrogate task to learn a robust feature extractor, which leads to the indirectly improved performance of density regressor. However, they did not consider the potential prediction perturbation on spatial regions due to the domain gap of feature learning from multi-tasks \cite{luo2021semi}; in which case, the binary segmentation can learn a different interest of the spatial regions compared with the one of the density regressor.
To address the challenge, we proposed a simple yet effective differential transformation layer \(\sigma(\cdot)\) to approximate the binary segmentation maps from the density regressor's output. In this way, we build a spatial regularization between the two tasks to address the potential inherent prediction perturbation issues in \cite{liu2020semi}. Meanwhile, the inherent consistency loss (\(L_{c{'}}\)) is employed between the binary segmentation predictions \((M_{B})\) and the approximated binary segmentation maps (\(M_{AB}\)) to regularize the surrogate task learning.
Note that, \(M_{B} \in \mathbb{R}^{H \times W \times 1}\) is the feature map of the corresponding crowd channel of the predicted class score \(P^{(H,W,C)} \in \mathbb{R}^{H \times W \times 2}\).

Following the same process that we generate the binary segmentation ground truth mask from the density map ground truth, to convert the predicted density maps into approximated binary segmentation maps, an intuitive way is to use the Heaviside step function to set all the positive pixel values in the predicted density maps to 1 and zero pixel values to 0. However, it is impractical in training because of the non-differentiability. Hence, we proposed a simple yet effective differential transformation function to guarantee that purpose. With the output from the density regressor \(M_{D}\), and the differential transformation layer \(\sigma(\cdot)\), the approximated binary segmentation map \(M_{AB}\) is defined as:
\begin{equation}
    M_{AB} = \sigma(K * M_{D}) = 2 * Sigmoid(K * M_{D}) -1,  
\end{equation}
where \(K\) is a very large constant, which is set as 6,000 in our work. Notably, as shown in Fig. \ref{network}, \(M_{D}\) is a non-negative density map prediction because of the use of ReLu as the activation.
In terms of such transformation function \(\sigma(\cdot)\), the spatial consistency can be enforced between the two different tasks in a trainable manner. Specifically, the density regressor focuses on the pixel values regression, while the binary segmentation predictor aims for semantic and spatial reasoning. Thus, the natural task-level prediction difference on spatial crowd regions of these two tasks can be regularized by an unsupervised inherent consistency loss function \(L_{c{'}}\) between the \(M_{B}\) and \(M_{AB}\).
\subsection{Loss Function}\label{sec:lossfunction}
We optimize the student model using the supervised loss (density regression, binary segmentation) on the labeled data and the unsupervised consistency loss on the unlabeled data. The whole network is end-to-end trainable, and the total loss function $L_{total}$ comprising five loss terms:

\begin{equation}
L_{total} = L_{Sd} + \alpha \cdot L_{Sb} +   L_{c{'}} +  \lambda \cdot (\alpha \cdot U_{h} \odot L_{Cb} + \cdot U_{s} \odot L_{Cd}),
\end{equation}
where \(\odot\) is Hadamard Product, $L_2$ loss is used for the supervised density map regression \(L_{Sd}\); categorical cross-entropy loss is used for supervised binary segmentation \(L_{Sb}\) in the student model. Besides, \(\alpha\) is a hyper-parameter to trade-off between the main task (density regression) and surrogate task (binary segmentation), which is set as 0.1 in our work. As for the unsupervised consistency loss, firstly, $L_2$ loss is used for unsupervised inherent consistency loss \(L_{c{'}}\) between the binary segmentation predictions and the approximated binary segmentation maps from density map predictions in the student model; secondly, 
`hard' uncertainty map \(U_{h}\) is used to assist the unsupervised consistency loss \(L_{Cb}\) upon the binary segmentation and `soft' uncertainty map \(U_{s}\) is used for unsupervised density map regression consistency loss \(L_{Cd}\). Here, we choose Euclidean distance as the consistency metric for \(L_{Cd}\) and \(L_{Cb}\). 
\(\lambda\) are adopted from \cite{laine2016temporal} as the same time-dependent Gaussian ramp-up weighting coefficient to trade-off between the supervised loss and unsupervised loss. 
This is to avoid the network get stuck in a degenerate solution, where no meaningful prediction of the unlabeled data is obtained \cite{laine2016temporal}.



\begin{figure*}
\centering
\includegraphics[width=17cm]{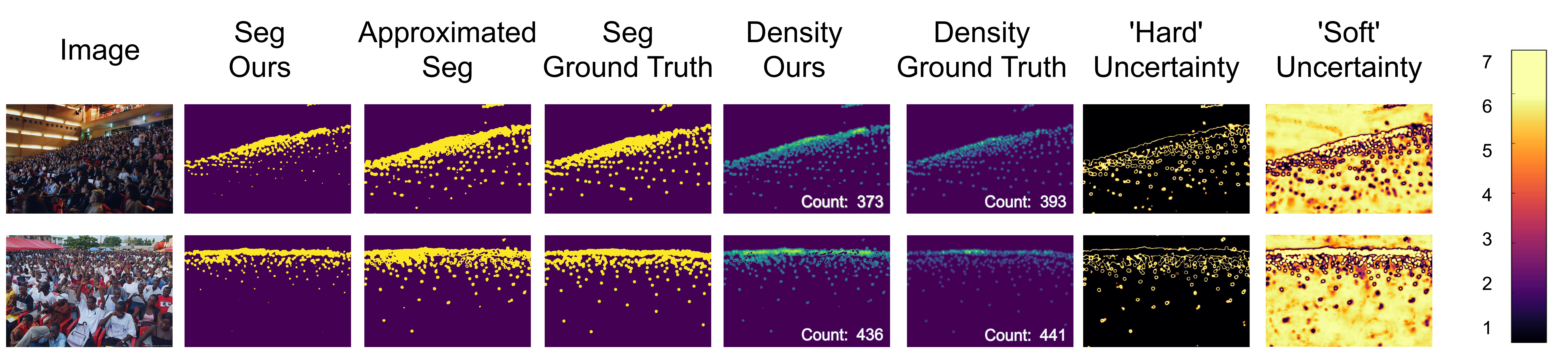}
\caption{Qualitative results on SHA test dataset. In the `hard' uncertainty maps, the yellow pixels represent uncertain regions and the black pixels are certain regions. In the `soft' uncertainty maps, the different color represents different weight mask values according to the color bar; higher value denotes more certain regions. The estimated `soft' uncertainty indicates that the crowd regions' boundary is more uncertain than other regions, which is reasonable because of the complex backgrounds. 
}
\label{allresults}
\end{figure*}

\begin{table*}[t]
\centering
\scalebox{0.8}
{
\begin{tabular}{c|c|cc|cc|cc|cc|cc}
\hline
\multicolumn{2}{c|}{\multirow{2}{*}{Methods}} & \multicolumn{2}{c|}{SHA} & \multicolumn{2}{c|}{SHB} & \multicolumn{2}{c|}{QNRF} & \multicolumn{2}{c}{JHU-Crowd} & \multicolumn{2}{c}{NWPU-Crowd} \\ \cline{3-12} 
\multicolumn{2}{c|}{} & \multicolumn{1}{c|}{MAE} & RMSE & \multicolumn{1}{c|}{MAE} & RMSE & \multicolumn{1}{c|}{MAE} & RMSE & \multicolumn{1}{c|}{MAE} & RMSE & MAE & RMSE \\ \hline
\multirow{3}{*}{Fully-Supervised}
& CACC \cite{liu2019context} & 62.3 & 100.0 & 7.8 & 12.2 & 107.0 & 183.0 & 100.1 & 314.0 & 93.6 & 489.9 \\
& CSR-Net \cite{li2018csrnet} & 68.2 & 115.0 & 10.6 & 16.0 & 119.2 & 211.4 & 85.9 & 309.2 & 104.9 & 433.5 \\
& Ours (Fully) & 66.9 & 125.6 & 12.3 & 17.9 & 119.2 & 213.3 & 80.1 & 305.3 & 105.8 & 445.3 \\\hline
\multirow{4}{*}{Semi-supervised} & Mean-Teacher \cite{tarvainen2017mean} (Baseline) & 88.2 & 151.1 & 15.9 & 25.7 & 147.2 & 249.6 & 121.5 & 388.9 & 129.8 & 515.0 \\
 & L2R \cite{liu2018leveraging} & 86.5 & 148.2 & 16.8 & 25.1 & 145.1 & 256.1 & 123.6 & 376.1 & 125.0 & 501.9 \\
 & Sindagi \textit{et al.} \cite{sindagi2020learning} & 89.0 & - & - & - & 136.0 & - & - & - & - & - \\
 & Liu \textit{et al.} \cite{liu2020semi} & - & - & - & - & 138.9 & - & - & - & - & - \\ \hline
\multicolumn{2}{c|}{Ours (Label-Only)} & 94.6 & 152.0 & 19.2 & 31.9 & 152.9 & 266.1 & 133.3 & 415.0 & 141.0 & 625.6 \\
\multicolumn{2}{c|}{\textbf{Ours (Semi)}} & \textbf{68.5} & \textbf{121.9} & \textbf{14.1} & \textbf{20.6} & \textbf{130.3} & \textbf{226.3} &\textbf{80.7}  & \textbf{290.8} &\textbf{111.7}  &\textbf{443.2} \\ \hline
\end{tabular}
}
\caption{Quantitative results on four crowd counting datasets. Our model achieves superior performance than the other semi-supervised methods in terms of MAE with the same setting of 50\% labeled data on four datasets.} 
\label{totalresult}
\end{table*}

\section{Experiments}
\subsection{Datasets}
\textbf{ShanghaiTech} \cite{zhang2016single} consists of 1,198 images, containing a total amount of 330,165 people with head centre point annotations. This dataset has two parts: \textbf{SHA} includes 482 images and is divided into a training (300) and testing (182) subset. \textbf{SHB} includes 716 images and is divided into 400 images for training and 316 images for testing.  
\textbf{UCF-QNRF} \cite{idrees2018composition} is a large crowd dataset, consisting of 1,535 images with about 1.25 million annotations in total.
As indicated by \cite{idrees2018composition}, 1,201 images are used for training; the remaining 334 images form the test set. 
\textbf{JHU-Crowd} \cite{sindagi2019pushing} is a recent challenging large-scale dataset that containing 4,372 images with 1.51 million annotations. 
This dataset is divided into 2,272 images for training, 500 images for validation, and 1,600 images for testing. 
\textbf{NWPU-Crowd} \cite{gao2020nwpu} is current the largest public crowd counting dataset, containing 5,109 images with over 2.13 million annotations. The dataset includes 3109 training images and 500 validation images; due to no access to the testing images; instead, we keep their validation images to evaluate our model's performance.
Note that, we set 50\% of the training images as the labeled data and the rest as the unlabeled data. In particular, for ShanghaiTech (part A, part B), UCF-QNRF and NWPU-Crowd, we use 10\% of the labeled training images as the validation dataset.

\subsection{Implementation Details}
We adopt 
a truncated VGG-16 \cite{Simonyan15} as the backbone network, which is the same as \cite{liu2019context,li2018csrnet,liu2020semi,tarvainen2017mean,liu2018leveraging}.
Additionally, following \cite{tarvainen2017mean}, two dropout layers with a drop out rate of 0.5 are added into the feature extractor to introduce model-level perturbations. The dropout is turned on during the training and turned off during the testing. Please refer to supplementary for detailed model structure. We update the teacher model's weight \(\theta{'}\) as an EMA of the student model's weight \(\theta\) during the training step, such as \(\theta{'}_{t} = \zeta \cdot \theta{'}_{t-1} + (1 - \zeta) \cdot \theta_{t}\), where \(t\) is the \(t_{th}\) training step, and \(\zeta\) is the EMA decay to control the update rate, which is empirically set as 0.999 in our work. For Shannon Entropy estimation, we set \(T = 8\) as the stochastic forward passes times to balance the model's performance and training efficiency. Besides, we set the \textit{threshold} as a Gaussian ramp-up function from 3/4 maximum uncertainty value to maximum uncertainty value for `hard' uncertainty map estimation. For the `soft' uncertainty map estimation, the weight value \(M\) is set as 7. Details of the hyper-parameter setting in our work can be found in the supplement. 

The training data set is augmented by randomly cropping the input images, the density maps ground truth, and the binary segmentation ground truth with fixed size \(128 \times 128\) at a random location; then randomly horizontal flipped the image patches with the probability of 0.3. 
We trained our model up to 600 epochs or stop early when the network has converged, with an initial learning rate of 7e-5 and divided by 5 every 200 epochs. The batch size is set as 16, consisting of 8 labeled images and 8 unlabeled images. All the training processes are performed on a server with 8 TESLA V100, and all the testing experiments are conducted on a local workstation with a Geforce RTX 2080Ti.


\section{Results}
In this section, we present our experimental results on the crowd counting tasks compared to previous state-of-the-art methods. Following the previous methods, we adopt Mean Absolute Error \((MAE)\) and Root Mean Squared Error \((RMSE)\) to evaluate the counting performance.
The results of ablation study are also shown to demonstrate the importance of the various components in our framework, such as the number of labeled and unlabeled images, `soft' and `hard' uncertainty maps, differential transformation layer, respectively.
Quantitative results are shown in  Tab. \ref{totalresult}, \ref{unc} and Fig. \ref{numlabeled}. Fig. \ref{allresults} shows the qualitative results. More qualitative results can be found in the supplementary.  More quantitative results compared with previous methods (\cite{liu2020semi,sindagi2020learning,tarvainen2017mean}) under different number of labeled data settings are shown in the supplementary.
\subsection{Crowd Counting Results }
Fig. \ref{allresults} shows qualitative results; specifically, we present the predicted and approximated segmentation maps, and the visualized uncertainty maps to demonstrate our model's cohesion, along with the contribution of spatial uncertainty guidance and inherent consistency regularization. 
In particular, we compare our model with previous semi-supervised methods \cite{liu2018leveraging,liu2020semi,sindagi2020learning,tarvainen2017mean}. The results of \cite{liu2020semi,sindagi2020learning} are retrieved from their published papers, and we re-implement the rest methods \cite{liu2018leveraging,tarvainen2017mean} through running their public code.
Note that, \cite{liu2020semi} adopts the same backbone (VGG-16 \cite{Simonyan15}) as our model; they build their model based on CSRNet \cite{li2018csrnet}, which achieves a comparable performance under fully-supervised manner with ours (\textit{i.e. Ours (Fully) in Tab.  \ref{totalresult}}). \cite{sindagi2020learning} adopts a more powerful backbone producing superior performance than \textit{Ours (Fully)} under fully-supervised manner. So the comparison with them in a semi-supervised manner can be seen as straightforward and reasonable. 
Additionally, we add binary segmentation module into the Baseline model \cite{tarvainen2017mean} to maintain similar model parameters as Ours (Semi); however, without the proposed transformation layer and uncertainty maps, the Baseline model achieves relatively 18.5 \% higher MAE compared with Ours (Semi) on four datasets.
To make an intuitive comparison, we also present different prediction results with our proposed model: (1) Ours (Label-Only): trained with half labeled data on the student model (without transformation layer). (2) Ours (Semi): trained with half labeled and half unlabeled data on the student and teacher model simultaneously; inferred with student model only. (3) Ours (Fully): trained with all the labeled data on the student model (without transformation layer). 
Note that, 
the transformation layer works as an activation function, which hardly increases the size of the model.
Tab. \ref{totalresult} shows that Ours (Semi) outperforms the Ours (Label-Only) by a large margin with average 25.1\% performance gain in terms of MAE on four datasets, which is benefits from the proposed uncertainty maps, differential transformation layer and unlabeled data. In particular, our model achieves comparable performance with only 50\% labeled data, compared with Ours (Fully) with 100\% labeled data in SHA and JHU-Crowd dataset.
Furthermore, to present comprehensive comparisons, we also show the performance of previous state-of-the-art crowd counting methods \cite{li2018csrnet,liu2019context} with the same backbone network as ours under a fully supervised manner.
Tab. \ref{totalresult} shows our method outperforms other semi-supervised methods in terms of MAE and RMSE on all four datasets under the same test settings and achieves a comparable performance to the previous state-of-the-art fully supervised works in SHA and JHU-Crowd dataset.  




\subsection{Ablation Study}
We investigate the effect of each component in our proposed model. Our model is robust to the hyper-parameters; results of more ablation studies, such as coefficients of the loss function, \textit{threshold} of `hard' uncertainty map, weights of `soft' uncertainty map, \(etc.\), can be found in the supplementary.

\begin{figure}[th]
\centering
\includegraphics[width=4cm]{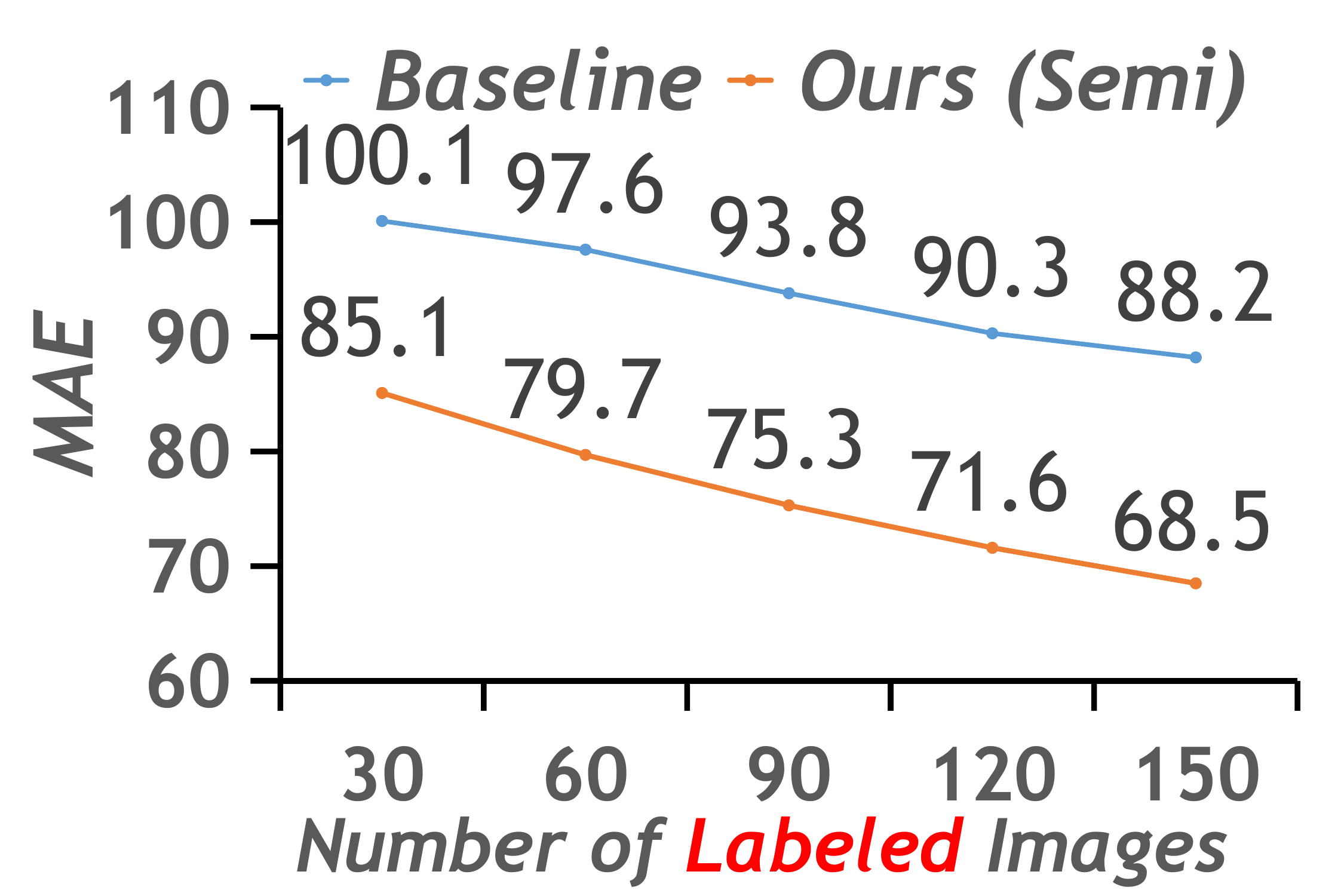}
\includegraphics[width=4cm]{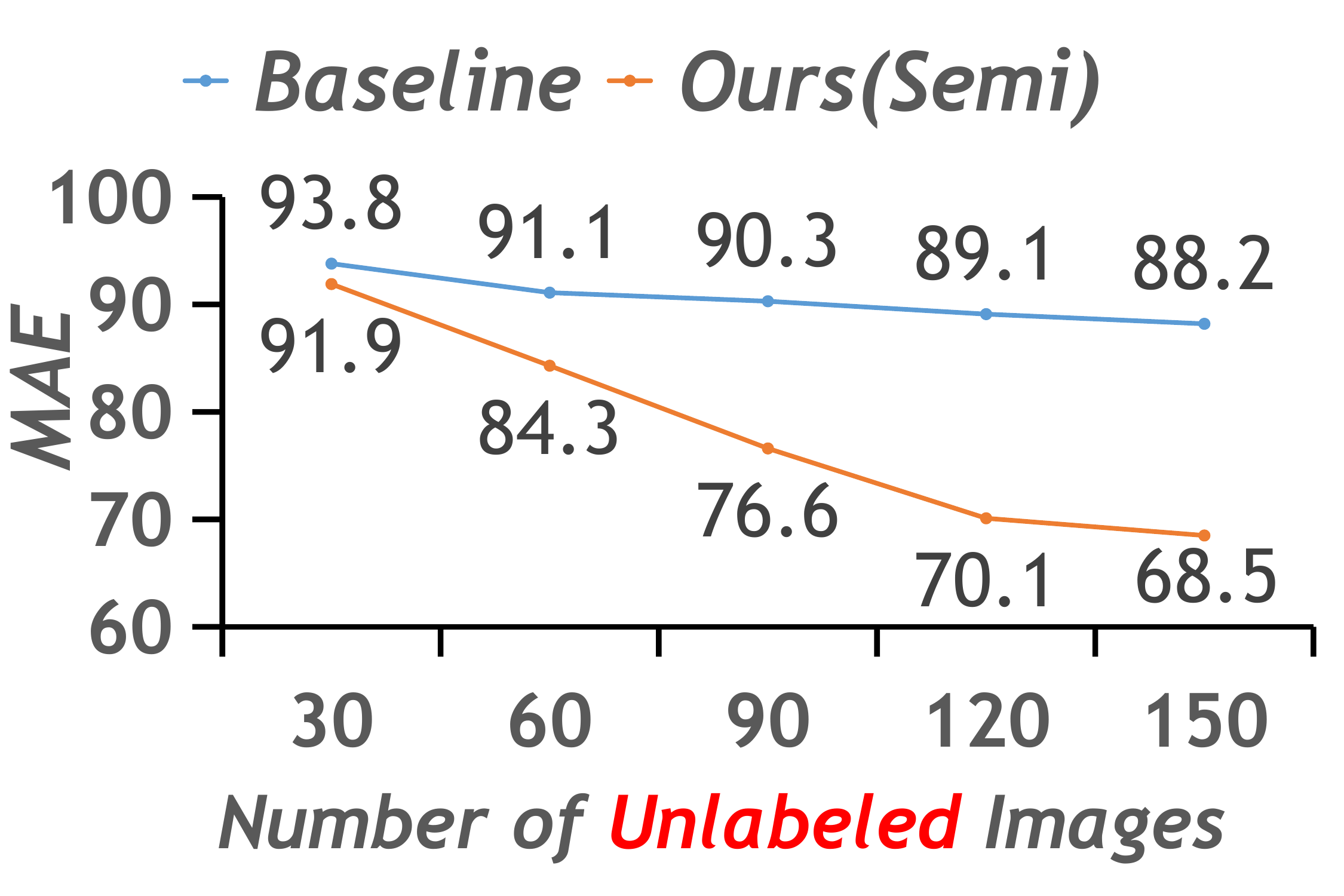}
\caption{The impact of the number of labeled \& unlabeled images. Evaluated on SHA dataset in terms of MAE.}
\label{numlabeled}
\end{figure}

\textbf{Ablation on Number of Labeled \& Unlabeled Images:}
We examine the performance of Baseline \cite{tarvainen2017mean} and Ours (Semi) with a different number of labeled \& unlabeled images. We conduct experiments on the SHA dataset by varying the number of labeled images from 30 to 150 while fixing the number of unlabeled images to be 150; or varying the number of unlabeled images from 30 to 150 while fixing the amount of labeled images to be 150. The performance are shown in Fig. \ref{numlabeled}, where it shows Ours (Semi) achieves consistent superior performance over the Baseline \cite{tarvainen2017mean}, which demonstrate the robustness of our method.

\textbf{Ablation on Uncertainty Map:}
We conduct several experiments to evaluate the impact of the proposed uncertainty maps (Unc). 
Firstly, we remove both the `hard' and `soft' uncertainty maps and keep the rest model structure. Notably, the concept of 'surrogate task' is used for spatial uncertainty estimation from binary segmentation task in this work; if we remove the uncertainty module, the binary segmentation task will only be served as information supplement for intermediate feature learning.
Secondly, we add either `hard' uncertainty map or `soft' uncertainty map respectively to evaluate the effectiveness of each of them.
Thirdly, we add two `hard' uncertainty maps to verify the effectiveness of the proposed `soft' uncertainty map with respect to the consistency learning on the density regression. 
Finally, we add both `hard' and `soft' uncertainty maps (Ours) for further comparison. 
Tab. \ref{unc} shows that our model with both uncertainty maps achieves average 15.5\% and 16.0\% performance gain via MAE compared with that without uncertainty map employed on SHA and JHU-Crowd datasets, respectively. This proves that our proposed uncertainty maps can assist the feature learning between the student and teacher model and further improve the performance.

\begin{table}[th]
\centering
\scalebox{0.8}
{
\begin{tabular}{c|c|c|c|c}
\hline
\multirow{2}{*}{Methods} & \multicolumn{2}{c|}{SHA} & \multicolumn{2}{c}{JHU-Crowd} \\ \cline{2-5} 
 & MAE & RMSE & MAE & RMSE \\ \hline
w/o Unc & 81.1 & 143.1 & 96.1 & 311.9 \\
w/ `Hard' Unc & 77.3 & 137.0 & 92.7 & 304.0 \\
w/ `Soft' Unc & 73.1 & 130.8 & 85.3 & 296.2 \\ 
w/ two `Soft' Unc & 70.5  & 124.6 & 83.9 & 295.2 \\
w/ two `Hard' Unc & 72.1  & 128.9 & 83.2 & 294.8 \\\hline
\textbf{w/ both Unc (ours)} & \textbf{68.5} & \textbf{121.9} & \textbf{80.7} & \textbf{290.8} \\ \hline
\end{tabular}
}
\caption{Performance comparison of the effectiveness of the proposed uncertainty maps. Compared with the `hard' uncertainty maps, the `soft' uncertainty maps can bring average 6.5\% superior performance improvement via MAE on two datasets.}
\label{unc}
\end{table}


\section{Conclusions}
We propose a spatial uncertainty-aware semi-supervised crowd counting methodology via regularized surrogate task to alleviate the inevitable noisy supervision from the unlabeled data. We have demonstrated its potentials in reducing annotations efforts while maintaining good performance upon four challenging crowd counting datasets.
It is anticipated that our approach will be widely applicable in the real world.

\noindent\textbf{Acknowledgements}
Y. Meng and H. Zhang thank the China Science IntelliCloud Technology Co., Ltd for the studentships We thank NVIDIA for the donation of GPU cards. This work was undertaken on Barkla, part of the High Performance Computing facilities at the University of Liverpool, Liverpool, United Kingdom.
{\small
\bibliographystyle{ieee_fullname}
\bibliography{egbib_final}

\begin{thebibliography}{10}\itemsep=-1pt

\bibitem{bai2020adaptive}
Shuai Bai, Zhiqun He, Yu Qiao, Hanzhe Hu, Wei Wu, and Junjie Yan.
\newblock Adaptive dilated network with self-correction supervision for
  counting.
\newblock In {\em Proceedings of the IEEE Conference on Computer Vision and
  Pattern Recognition}, pages 4594--4603, 2020.

\bibitem{baumgartner2019phiseg}
Christian~F Baumgartner, Kerem~C Tezcan, Krishna Chaitanya, Andreas~M
  H{\"o}tker, Urs~J Muehlematter, Khoschy Schawkat, Anton~S Becker, Olivio
  Donati, and Ender Konukoglu.
\newblock Phiseg: Capturing uncertainty in medical image segmentation.
\newblock In {\em International Conference on Medical Image Computing and
  Computer-Assisted Intervention}, pages 119--127. Springer, 2019.

\bibitem{boominathan2016crowdnet}
Lokesh Boominathan, Srinivas~SS Kruthiventi, and R~Venkatesh Babu.
\newblock Crowdnet: A deep convolutional network for dense crowd counting.
\newblock In {\em Proceedings of the 24th ACM international conference on
  Multimedia}, pages 640--644, 2016.

\bibitem{bridge2020introducing}
Joshua Bridge, Yanda Meng, Yitian Zhao, Yong Du, Mingfeng Zhao, Renrong Sun,
  and Yalin Zheng.
\newblock Introducing the {GEV} activation function for highly unbalanced data
  to develop {COVID-19} diagnostic models.
\newblock {\em IEEE Journal of Biomedical and Health Informatics},
  24(10):2776--2786, 2020.

\bibitem{chen2021unsupervised}
Xu Chen, Yanda Meng, Yitian Zhao, Rachel Williams, R.~Vallabhaneni Srinivasa,
  and Yalin Zheng.
\newblock Learning parameter-specific affine transformation for medical images
  registration.
\newblock In {\em International Conference on Medical Image Computing and
  Computer-Assisted Intervention}, 2021.

\bibitem{chen2019learning}
Xu Chen, Bryan~M Williams, Srinivasa~R Vallabhaneni, Gabriela Czanner, Rachel
  Williams, and Yalin Zheng.
\newblock Learning active contour models for medical image segmentation.
\newblock In {\em Proceedings of the IEEE/CVF Conference on Computer Vision and
  Pattern Recognition}, pages 11632--11640, 2019.

\bibitem{gao2019pcc}
Junyu Gao, Qi Wang, and Xuelong Li.
\newblock Pcc net: Perspective crowd counting via spatial convolutional
  network.
\newblock {\em IEEE Transactions on Circuits and Systems for Video Technology},
  2019.

\bibitem{he2019bounding}
Yihui He, Chenchen Zhu, Jianren Wang, Marios Savvides, and Xiangyu Zhang.
\newblock Bounding box regression with uncertainty for accurate object
  detection.
\newblock In {\em Proceedings of the IEEE Conference on Computer Vision and
  Pattern Recognition}, pages 2888--2897, 2019.

\bibitem{idrees2018composition}
Haroon Idrees, Muhmmad Tayyab, Kishan Athrey, Dong Zhang, Somaya Al-Maadeed,
  Nasir Rajpoot, and Mubarak Shah.
\newblock Composition loss for counting, density map estimation and
  localization in dense crowds.
\newblock In {\em Proceedings of the European Conference on Computer Vision
  (ECCV)}, pages 532--546, 2018.

\bibitem{jiang2019crowd}
Xiaolong Jiang, Zehao Xiao, Baochang Zhang, Xiantong Zhen, Xianbin Cao, David
  Doermann, and Ling Shao.
\newblock Crowd counting and density estimation by trellis encoder-decoder
  networks.
\newblock In {\em Proceedings of the IEEE Conference on Computer Vision and
  Pattern Recognition}, pages 6133--6142, 2019.

\bibitem{jiang2020attention}
Xiaoheng Jiang, Li Zhang, Mingliang Xu, Tianzhu Zhang, Pei Lv, Bing Zhou, Xin
  Yang, and Yanwei Pang.
\newblock Attention scaling for crowd counting.
\newblock In {\em Proceedings of the IEEE Conference on Computer Vision and
  Pattern Recognition}, pages 4706--4715, 2020.

\bibitem{kendall2017uncertainties}
Alex Kendall and Yarin Gal.
\newblock What uncertainties do we need in bayesian deep learning for computer
  vision?
\newblock In {\em Advances in neural information processing systems}, pages
  5574--5584, 2017.

\bibitem{kohl2018probabilistic}
Simon Kohl, Bernardino Romera-Paredes, Clemens Meyer, Jeffrey De~Fauw, Joseph~R
  Ledsam, Klaus Maier-Hein, SM~Ali Eslami, Danilo~Jimenez Rezende, and Olaf
  Ronneberger.
\newblock A probabilistic u-net for segmentation of ambiguous images.
\newblock In {\em Advances in Neural Information Processing Systems}, pages
  6965--6975, 2018.

\bibitem{laine2016temporal}
Samuli Laine and Timo Aila.
\newblock Temporal ensembling for semi-supervised learning.
\newblock {\em arXiv preprint arXiv:1610.02242}, 2016.

\bibitem{lempitsky2010learning}
Victor Lempitsky and Andrew Zisserman.
\newblock Learning to count objects in images.
\newblock In {\em Advances in neural information processing systems}, pages
  1324--1332, 2010.

\bibitem{li2018csrnet}
Yuhong Li, Xiaofan Zhang, and Deming Chen.
\newblock Csrnet: Dilated convolutional neural networks for understanding the
  highly congested scenes.
\newblock In {\em Proceedings of the IEEE conference on computer vision and
  pattern recognition}, pages 1091--1100, 2018.

\bibitem{lian2019density}
Dongze Lian, Jing Li, Jia Zheng, Weixin Luo, and Shenghua Gao.
\newblock Density map regression guided detection network for rgb-d crowd
  counting and localization.
\newblock In {\em Proceedings of the IEEE Conference on Computer Vision and
  Pattern Recognition}, pages 1821--1830, 2019.

\bibitem{liu2019recurrent}
Chenchen Liu, Xinyu Weng, and Yadong Mu.
\newblock Recurrent attentive zooming for joint crowd counting and precise
  localization.
\newblock In {\em 2019 IEEE/CVF Conference on Computer Vision and Pattern
  Recognition (CVPR)}, pages 1217--1226. IEEE, 2019.

\bibitem{liu2018decidenet}
Jiang Liu, Chenqiang Gao, Deyu Meng, and Alexander~G Hauptmann.
\newblock Decidenet: Counting varying density crowds through attention guided
  detection and density estimation.
\newblock In {\em Proceedings of the IEEE Conference on Computer Vision and
  Pattern Recognition}, pages 5197--5206, 2018.

\bibitem{liu2019adcrowdnet}
Ning Liu, Yongchao Long, Changqing Zou, Qun Niu, Li Pan, and Hefeng Wu.
\newblock Adcrowdnet: An attention-injective deformable convolutional network
  for crowd understanding.
\newblock In {\em Proceedings of the IEEE Conference on Computer Vision and
  Pattern Recognition}, pages 3225--3234, 2019.

\bibitem{liu2019context}
Weizhe Liu, Mathieu Salzmann, and Pascal Fua.
\newblock Context-aware crowd counting.
\newblock In {\em Proceedings of the IEEE Conference on Computer Vision and
  Pattern Recognition}, pages 5099--5108, 2019.

\bibitem{liu2018leveraging}
Xialei Liu, Joost Van De~Weijer, and Andrew~D Bagdanov.
\newblock Leveraging unlabeled data for crowd counting by learning to rank.
\newblock In {\em Proceedings of the IEEE Conference on Computer Vision and
  Pattern Recognition}, pages 7661--7669, 2018.

\bibitem{liu2019exploiting}
Xialei Liu, Joost Van De~Weijer, and Andrew~D Bagdanov.
\newblock Exploiting unlabeled data in cnns by self-supervised learning to
  rank.
\newblock {\em IEEE transactions on pattern analysis and machine intelligence},
  41(8):1862--1878, 2019.

\bibitem{liu2020semi}
Yan Liu, Lingqiao Liu, Peng Wang, Pingping Zhang, and Yinjie Lei.
\newblock Semi-supervised crowd counting via self-training on surrogate tasks.
\newblock {\em European Conference on Computer Vision}, 2020.

\bibitem{liu2019point}
Yuting Liu, Miaojing Shi, Qijun Zhao, and Xiaofang Wang.
\newblock Point in, box out: Beyond counting persons in crowds.
\newblock In {\em Proceedings of the IEEE Conference on Computer Vision and
  Pattern Recognition}, pages 6469--6478, 2019.

\bibitem{long2018transferable}
Mingsheng Long, Yue Cao, Zhangjie Cao, Jianmin Wang, and Michael~I Jordan.
\newblock Transferable representation learning with deep adaptation networks.
\newblock {\em IEEE transactions on pattern analysis and machine intelligence},
  41(12):3071--3085, 2018.

\bibitem{luo2019hybrid}
Ao Luo, Fan Yang, Xin Li, Dong Nie, Zhicheng Jiao, Shangchen Zhou, and Hong
  Cheng.
\newblock Hybrid graph neural networks for crowd counting.
\newblock {\em Proceedings of the AAAI Conference on Artificial Intelligence},
  2019.

\bibitem{luo2021semi}
Xiangde Luo, Jieneng Chen, Tao Song, Yinan Chen, Guotai Wang, and Shaoting
  Zhang.
\newblock Semi-supervised medical image segmentation through dual-task
  consistency.
\newblock {\em AAAI Conference on Artificial Intelligence}, 2021.

\bibitem{meng2020regression}
Yanda Meng, Wei Meng, Dongxu Gao, Yitian Zhao, Xiaoyun Yang, Xiaowei Huang, and
  Yalin Zheng.
\newblock Regression of instance boundary by aggregated cnn and gcn.
\newblock In {\em European Conference on Computer Vision}, pages 190--207.
  Springer, 2020.

\bibitem{meng2020cnn}
Yanda Meng, Meng Wei, Dongxu Gao, Yitian Zhao, Xiaoyun Yang, Xiaowei Huang, and
  Yalin Zheng.
\newblock Cnn-gcn aggregation enabled boundary regression for biomedical image
  segmentation.
\newblock In {\em International Conference on Medical Image Computing and
  Computer-Assisted Intervention}, pages 352--362. Springer, 2020.

\bibitem{miao2020shallow}
Yunqi Miao, Zijia Lin, Guiguang Ding, and Jungong Han.
\newblock Shallow feature based dense attention network for crowd counting.
\newblock In {\em Proceedings of the AAAI Conference on Artificial
  Intelligence}, pages 11765--11772, 2020.

\bibitem{oh2020crowd}
Min-hwan Oh, Peder~A Olsen, and Karthikeyan~Natesan Ramamurthy.
\newblock Crowd counting with decomposed uncertainty.
\newblock In {\em Proceedings of the AAAI Conference on Artificial
  Intelligence}, pages 11799--11806, 2020.

\bibitem{ranjan2020uncertainty}
Viresh Ranjan, Boyu Wang, Mubarak Shah, and Minh Hoai.
\newblock Uncertainty estimation and sample selection for crowd counting.
\newblock In {\em Proceedings of the Asian Conference on Computer Vision},
  2020.

\bibitem{sam2020locate}
Deepak~Babu Sam, Skand~Vishwanath Peri, Mukuntha~Narayanan Sundararaman, Amogh
  Kamath, and Venkatesh~Babu Radhakrishnan.
\newblock Locate, size and count: Accurately resolving people in dense crowds
  via detection.
\newblock {\em IEEE Transactions on Pattern Analysis and Machine Intelligence},
  2020.

\bibitem{sam2019almost}
Deepak~Babu Sam, Neeraj~N Sajjan, Himanshu Maurya, and R~Venkatesh Babu.
\newblock Almost unsupervised learning for dense crowd counting.
\newblock In {\em Proceedings of the AAAI Conference on Artificial
  Intelligence}, volume~33, pages 8868--8875, 2019.

\bibitem{shannon2001mathematical}
Claude~Elwood Shannon.
\newblock A mathematical theory of communication.
\newblock {\em ACM SIGMOBILE mobile computing and communications review},
  5(1):3--55, 2001.

\bibitem{shannon1949mathematical}
Claude~Elwood Shannon and Warren Weaver.
\newblock The mathematical theory of communication. univ.
\newblock {\em Illinois press, Urbana, I}, 11:117, 1949.

\bibitem{shen2018crowd}
Zan Shen, Yi Xu, Bingbing Ni, Minsi Wang, Jianguo Hu, and Xiaokang Yang.
\newblock Crowd counting via adversarial cross-scale consistency pursuit.
\newblock In {\em Proceedings of the IEEE conference on computer vision and
  pattern recognition}, pages 5245--5254, 2018.

\bibitem{shi2019revisiting}
Miaojing Shi, Zhaohui Yang, Chao Xu, and Qijun Chen.
\newblock Revisiting perspective information for efficient crowd counting.
\newblock In {\em Proceedings of the IEEE Conference on Computer Vision and
  Pattern Recognition}, pages 7279--7288, 2019.

\bibitem{shi2019counting}
Zenglin Shi, Pascal Mettes, and Cees~GM Snoek.
\newblock Counting with focus for free.
\newblock In {\em Proceedings of the IEEE International Conference on Computer
  Vision}, pages 4200--4209, 2019.

\bibitem{shi2018crowd}
Zenglin Shi, Le Zhang, Yun Liu, Xiaofeng Cao, Yangdong Ye, Ming-Ming Cheng, and
  Guoyan Zheng.
\newblock Crowd counting with deep negative correlation learning.
\newblock In {\em Proceedings of the IEEE conference on computer vision and
  pattern recognition}, pages 5382--5390, 2018.

\bibitem{Simonyan15}
Karen Simonyan and Andrew Zisserman.
\newblock Very deep convolutional networks for large-scale image recognition.
\newblock In {\em International Conference on Learning Representations}, 2015.

\bibitem{sindagi2019ha}
Vishwanath~A Sindagi and Vishal~M Patel.
\newblock Ha-ccn: Hierarchical attention-based crowd counting network.
\newblock {\em IEEE Transactions on Image Processing}, 29:323--335, 2019.

\bibitem{sindagi2019multi}
Vishwanath~A Sindagi and Vishal~M Patel.
\newblock Multi-level bottom-top and top-bottom feature fusion for crowd
  counting.
\newblock In {\em Proceedings of the IEEE International Conference on Computer
  Vision}, pages 1002--1012, 2019.

\bibitem{sindagi2020learning}
Vishwanath~A Sindagi, Rajeev Yasarla, Deepak~Sam Babu, R~Venkatesh Babu, and
  Vishal~M Patel.
\newblock Learning to count in the crowd from limited labeled data.
\newblock {\em European Conference on Computer Vision}, 2020.

\bibitem{sindagi2019pushing}
Vishwanath~A Sindagi, Rajeev Yasarla, and Vishal~M Patel.
\newblock Pushing the frontiers of unconstrained crowd counting: New dataset
  and benchmark method.
\newblock In {\em Proceedings of the IEEE International Conference on Computer
  Vision}, pages 1221--1231, 2019.

\bibitem{sindagi2020jhu}
Vishwanath~A Sindagi, Rajeev Yasarla, and Vishal~M Patel.
\newblock Jhu-crowd++: Large-scale crowd counting dataset and a benchmark
  method.
\newblock {\em arXiv preprint arXiv:2004.03597}, 2020.

\bibitem{srivastava2014dropout}
Nitish Srivastava, Geoffrey Hinton, Alex Krizhevsky, Ilya Sutskever, and Ruslan
  Salakhutdinov.
\newblock Dropout: a simple way to prevent neural networks from overfitting.
\newblock {\em The journal of machine learning research}, 15(1):1929--1958,
  2014.

\bibitem{tarvainen2017mean}
Antti Tarvainen and Harri Valpola.
\newblock Mean teachers are better role models: Weight-averaged consistency
  targets improve semi-supervised deep learning results.
\newblock In {\em Advances in neural information processing systems}, pages
  1195--1204, 2017.

\bibitem{von2016gaussian}
Matthias von Borstel, Melih Kandemir, Philip Schmidt, Madhavi~K Rao, Kumar
  Rajamani, and Fred~A Hamprecht.
\newblock Gaussian process density counting from weak supervision.
\newblock In {\em European Conference on Computer Vision}, pages 365--380.
  Springer, 2016.

\bibitem{vu2019advent}
Tuan-Hung Vu, Himalaya Jain, Maxime Bucher, Matthieu Cord, and Patrick
  P{\'e}rez.
\newblock Advent: Adversarial entropy minimization for domain adaptation in
  semantic segmentation.
\newblock In {\em Proceedings of the IEEE conference on computer vision and
  pattern recognition}, pages 2517--2526, 2019.

\bibitem{wan2019residual}
Jia Wan, Wenhan Luo, Baoyuan Wu, Antoni~B Chan, and Wei Liu.
\newblock Residual regression with semantic prior for crowd counting.
\newblock In {\em Proceedings of the IEEE Conference on Computer Vision and
  Pattern Recognition}, pages 4036--4045, 2019.

\bibitem{gao2020nwpu}
Qi Wang, Junyu Gao, Wei Lin, and Xuelong Li.
\newblock Nwpu-crowd: A large-scale benchmark for crowd counting and
  localization.
\newblock {\em IEEE Transactions on Pattern Analysis and Machine Intelligence},
  2020.

\bibitem{wang2019learning}
Qi Wang, Junyu Gao, Wei Lin, and Yuan Yuan.
\newblock Learning from synthetic data for crowd counting in the wild.
\newblock In {\em Proceedings of the IEEE conference on computer vision and
  pattern recognition}, pages 8198--8207, 2019.

\bibitem{wang2020pixel}
Qi Wang, Junyu Gao, Wei Lin, and Yuan Yuan.
\newblock Pixel-wise crowd understanding via synthetic data.
\newblock {\em International Journal of Computer Vision}, pages 1--21, 2020.

\bibitem{yang2020reverse}
Yifan Yang, Guorong Li, Zhe Wu, Li Su, Qingming Huang, and Nicu Sebe.
\newblock Reverse perspective network for perspective-aware object counting.
\newblock In {\em Proceedings of the IEEE/CVF Conference on Computer Vision and
  Pattern Recognition}, pages 4374--4383, 2020.

\bibitem{yangweakly2020}
Yifan Yang, Guorong Li, Zhe Wu, Li Su, Qingming Huang, and Nicu Sebe.
\newblock Weakly-supervised crowd counting learns from sorting rather than
  locations.
\newblock In {\em European Conference on Computer Vision}, 2020.

\bibitem{yu2019uncertainty}
Lequan Yu, Shujun Wang, Xiaomeng Li, Chi-Wing Fu, and Pheng-Ann Heng.
\newblock Uncertainty-aware self-ensembling model for semi-supervised 3d left
  atrium segmentation.
\newblock In {\em International Conference on Medical Image Computing and
  Computer-Assisted Intervention}, pages 605--613. Springer, 2019.

\bibitem{zhang2019attentional}
Anran Zhang, Lei Yue, Jiayi Shen, Fan Zhu, Xiantong Zhen, Xianbin Cao, and Ling
  Shao.
\newblock Attentional neural fields for crowd counting.
\newblock In {\em Proceedings of the IEEE International Conference on Computer
  Vision}, pages 5714--5723, 2019.

\bibitem{zhang2015cross}
Cong Zhang, Hongsheng Li, Xiaogang Wang, and Xiaokang Yang.
\newblock Cross-scene crowd counting via deep convolutional neural networks.
\newblock In {\em Proceedings of the IEEE conference on computer vision and
  pattern recognition}, pages 833--841, 2015.

\bibitem{zhang2021regularization}
Hongrun Zhang, Yanda Meng, Xuesheng Qian, Xiaoyun Yang, Sarah~E Coupland, and
  Yalin Zheng.
\newblock A regularization term for slide correlation reduction in whole slide
  image analysis with deep learning.
\newblock In {\em Medical Imaging with Deep Learning}, 2021.

\bibitem{Zhang2020UCNet}
Jing Zhang, Deng-Ping Fan, Yuchao Dai, Saeed Anwar, Fatemeh Sadat~Saleh, Tong
  Zhang, and Nick Barnes.
\newblock Uc-net: Uncertainty inspired rgb-d saliency detection via conditional
  variational autoencoders.
\newblock In {\em Proceedings of the IEEE conference on computer vision and
  pattern recognition}, 2020.

\bibitem{zhang2019wide}
Qi Zhang and Antoni~B Chan.
\newblock Wide-area crowd counting via ground-plane density maps and multi-view
  fusion cnns.
\newblock In {\em Proceedings of the IEEE Conference on Computer Vision and
  Pattern Recognition}, pages 8297--8306, 2019.

\bibitem{zhang2016single}
Yingying Zhang, Desen Zhou, Siqin Chen, Shenghua Gao, and Yi Ma.
\newblock Single-image crowd counting via multi-column convolutional neural
  network.
\newblock In {\em Proceedings of the IEEE conference on computer vision and
  pattern recognition}, pages 589--597, 2016.

\bibitem{zhao2019leveraging}
Muming Zhao, Jian Zhang, Chongyang Zhang, and Wenjun Zhang.
\newblock Leveraging heterogeneous auxiliary tasks to assist crowd counting.
\newblock In {\em Proceedings of the IEEE Conference on Computer Vision and
  Pattern Recognition}, pages 12736--12745, 2019.

\bibitem{zhao2015automated}
Yitian Zhao, Lavdie Rada, Ke Chen, Simon~P Harding, and Yalin Zheng.
\newblock Automated vessel segmentation using infinite perimeter active contour
  model with hybrid region information with application to retinal images.
\newblock {\em IEEE Transactions on Medical Imaging}, 34(9):1797--1807, 2015.

\bibitem{zhao2020active}
Zhen Zhao, Miaojing Shi, Xiaoxiao Zhao, and Li Li.
\newblock Active crowd counting with limited supervision.
\newblock {\em European Conference on Computer Vision}, 2020.

\bibitem{zou2019confidence}
Yang Zou, Zhiding Yu, Xiaofeng Liu, BVK Kumar, and Jinsong Wang.
\newblock Confidence regularized self-training.
\newblock In {\em Proceedings of the IEEE/CVF International Conference on
  Computer Vision}, pages 5982--5991, 2019.

\end{thebibliography}
}

\end{document}